\documentclass[conference]{IEEEtran}
\IEEEoverridecommandlockouts
\usepackage{amsmath,amssymb,amsfonts}
\usepackage{booktabs}
\usepackage{array}
\usepackage{multirow}
\usepackage{graphicx}
\usepackage{pgfplots}
\pgfplotsset{compat=1.17}
\usepgfplotslibrary{groupplots}
\usetikzlibrary{positioning,arrows.meta,fit,calc}
\definecolor{figblue}{HTML}{2A78D6}
\definecolor{figorange}{HTML}{EB6834}
\definecolor{figgray}{HTML}{898781}
\usepackage{cite}
\usepackage{url}
\usepackage{xcolor}

\begin{document}

\title{From Prompt to Variable: How Acceptance, Section, and Label Rules Shape LLM-Extracted Financial Disclosures\thanks{This work was supported by the PNC Presidential Fellowship Research Award that I hold, which funded much of the computing used here. The views expressed are my own and not necessarily those of PNC. Parts of this research ran on ORCHARD, a high-performance cloud computing cluster, and I thank Carnegie Mellon University for making it available to me.}}

\author{\IEEEauthorblockN{A. Orhun Gun}
\IEEEauthorblockA{Accounting AI Research Lab \\
Carnegie Mellon University \\
Pittsburgh, PA, USA \\
orhun@cmu.edu}}

\maketitle

\begin{abstract}
Two researchers using identical large language models (LLMs), prompts, and source documents can construct different financial variables and reach opposing empirical conclusions. Using 15,600 SEC Form 10-K filings (2000--2025) processed by four LLM configurations, I hold stored responses fixed and replay them under alternative acceptance, section, and label rules. First, all-or-nothing acceptance discards passing shock outputs whenever a companion output fails verification, favoring shorter, more recent disclosures. Separate acceptance raises four-configuration common-sample coverage for current/past shocks from 77.6\% to 98.6\% and reduces the width of Manski bounds on recorded shock incidence among substantive sections from 25.0 to 1.6 percentage points. Second, adding Business and Management's Discussion and Analysis to Risk Factors and counting a residual macroeconomic label increases measured current/past shock incidence from 33.7\% to 89.4\% in one configuration. Alternative label rules identify different sets of newly exposed filings even when aggregate increments nearly match. Together, section and label choices reverse covenant-distress regression coefficients across all four configurations, with detectable changes in two. Blind audits by two LLM judges, not validated against human coders, find no detectable loss in strict validity among recovered or added records, but consistently lower validity for residual labels. I provide a dataset release checklist and show why reproducible LLM-based measurement requires specifying acceptance rules, source sections, and counted labels alongside models and prompts.

\end{abstract}


\section{Introduction}
Large language models (LLMs) turn massive amounts of structured and unstructured data such as annual reports, conference calls, and news into financial variables~\cite{kim2023, dolphin2026, lopezlira2026}. The rules a pipeline applies before and after the LLM answers help define the variable a study uses. Researchers should treat these rules as part of variable construction, just as they treat other methodological choices in empirical finance. Finance and economics already distinguish missing R\&D from reported zeros~\cite{koh2015} and recognize the consequences of omitted delisting returns~\cite{shumway1997} and nonrandom sample selection~\cite{heckman1979}. Three such rules stand out: which outputs to accept, which sections to send, and which labels to count.

An all-or-nothing acceptance rule discards shock outputs that pass their own checks whenever a companion output fails. Consider a stored response to Skyworks Solutions' Risk Factors section (Item 1A). In it, the LLM reports one current/past shock (a macro or sector event affecting the firm now or earlier) and no future shocks (events the firm expects to affect it), and both outputs pass their own checks. Four of the response's 32 risk-factor quotations, split across bullet points, fail verification, so the pipeline rejects the whole response. Data and LLM validation tools also offer this all-or-nothing design~\cite{databricks2026, shankar2024}. For a researcher measuring corporate exposure to economic shocks from annual filings, the lost shock outputs are missing data, not evidence that the firm reported no shock. Coding them as zeros would understate current/past shock incidence.

A section rule decides where in the filing the extractor looks. Reading the whole 10-K costs more, and models use long contexts unevenly~\cite{liu2024}. Researchers send the section that SEC rules assign to the content, such as Item 1A for risk factors~\cite{sec10k, dolphin2026}. However, Unified Western Grocers discusses Sarbanes--Oxley (SOX) compliance in its Risk Factors but reports a related \$1.3M expense in Management's Discussion and Analysis (MD\&A, Item~7). A label rule decides whether the residual \texttt{other\_macro} category counts alongside named events. The same filing shows this rule too (Section~\ref{sec:labelrule}).

I built an extraction pipeline for finance research and stored its responses from four LLM configurations. I hold the responses fixed and replay them under different deterministic rules, so only the rules vary (Fig.~\ref{fig:schematic}). Under joint acceptance, one failed output rejects the whole response. Under separate acceptance, each output stands or falls on its own unchanged checks. Stage 1 compares the acceptance rules on all available Risk Factors sections. Stage 2 compares the section and label rules on the filings whose Business (Item 1), Risk Factors, and MD\&A sections passed joint acceptance in all four configurations. To see whether these rules reach regression estimates, I pair the shock variables with an LLM-extracted covenant distress flag from a companion extraction on the same filings~\cite{gun2026bl}.

\begin{figure}[t]
\centering
\begin{tikzpicture}[card/.style={draw=black!55, rounded corners=1pt, text width=.88\columnwidth, inner sep=3pt, align=left, font=\footnotesize, anchor=north west},
  flow/.style={-{Stealth[length=3pt]}, thin}]
\node[card, fill=black!5] (src) at (0,0) {\textbf{Stored responses.} For each section, the LLM returns three outputs: risk factors, current/past shocks, and future shocks.};
\node[card, line width=0.6pt] (one) at ([yshift=-5pt]src.south west) {\textbf{Stage 1 varies the acceptance rule in Item 1A.} At Skyworks Solutions, four of 32 risk-factor quotations fail. Joint acceptance leaves both shock outputs missing. Separate acceptance gives 1 for current/past shocks and 0 for future shocks.};
\node[card, line width=0.6pt] (two) at ([yshift=-5pt]one.south west) {\textbf{Stage 2 varies sections and labels (joint fixed).} At Unified Western Grocers, one configuration finds only the residual current/past label in Item 1A. Item 7 reports a \$1.3M Sarbanes--Oxley expense. Adding Items 1 and 7 moves the named flag from 0 to 1. The broad flag, which also counts the residual label, stays at 1.};
\coordinate (fork) at ([xshift=-5pt]src.west);
\draw[flow] (src.west) -- (fork) |- (two.west);
\draw[flow] (fork |- one.west) -- (one.west);
\end{tikzpicture}
\caption{I replay the stored responses in two comparisons: 12,398 Item 1A sections in Stage 1, and the 8,987 filings that all four configurations completed in Items 1, 1A, and 7 in Stage 2. A 1 flags a shock, 0 none.}
\label{fig:schematic}
\end{figure}

\begin{table}[!b]
\centering
\caption{What Each Rule Changes}
\label{tab:headline}
\scriptsize
\setlength{\tabcolsep}{2pt}
\begin{tabular*}{\columnwidth}{@{\extracolsep{\fill}}lrr@{}}
\toprule
\multicolumn{3}{@{}l}{\emph{Stage 1: acceptance rule, Item 1A current/past shocks}} \\
 & Joint & Separate \\
\midrule
Common sample (\% of 12,398) & 77.6 & 98.6 \\
2009 filings, any-configuration shock (\%) & 79.5 & 83.5 \\
\midrule
\multicolumn{3}{@{}l}{\emph{Stage 2: section and label rules, Gemma26 current/past shocks}} \\
Sections and labels & Flagged (\%) & \shortstack[r]{Distress on\\flag (points)} \\
\midrule
Item 1A, named & 33.7 & $-2.96$ (1.09) \\
Items 1, 1A, 7, named & 45.6 & $-2.06$ (1.02) \\
Item 1A, broad & 69.0 & $+0.02$ (1.18) \\
Items 1, 1A, 7, broad & 89.4 & $+5.28$ (1.39) \\
\bottomrule
\end{tabular*}
\par\vspace{2pt}
\parbox{\columnwidth}{\scriptsize The 2009 row omits placeholder Item 1A sections (under 1,000 characters). Broad also counts \texttt{other\_macro}. Stage 2 shows \texttt{Gemma26} (Section~\ref{sec:corpus}), the configuration with the widest swing (Table~\ref{tab:stage2_reg}), on the 8,987 panel filings. Regressions use the 8,864 filings with Standard Industrial Classification (SIC) codes, control for log Item 1A length, year, and industry effects, with firm-clustered standard errors (SE) in parentheses. From the first Stage 2 row to the last, the coefficient changes by 8.23 points (SE 1.74, from unrounded values).}
\end{table}

Three results stand out (Tables~\ref{tab:headline} and~\ref{tab:stage3_audit}). First, joint acceptance drops over a fifth of available Item 1A sections from the sample common to all four configurations, although each configuration alone loses under 12\%. What remains tilts toward short, recent sections. Separate acceptance adds back older, longer sections and, with all labels counted, raises the 2009 share of sections in which any configuration records a shock. Pooled regression coefficients change little (Table~\ref{tab:stage1_levels}, Panel B).

Second, the section and label rules together move the share of Stage~2 filings that one configuration flags for current/past shocks from $34\%$ to $89\%$. They also reverse the sign of that configuration's distress regression. Exposed filers look less distressed under the narrowest rules and more under the broadest.

Third, two blind LLM judges from different developers find that a weighted $25\%$ to $44\%$ of audited current/past records meet the full audit definition (strict validity). Both accept residual-label records less often than named ones. Neither detects a loss of strict validity in recovered Item 1A records or in Business and MD\&A additions.

\section{Related Work}
\textbf{Text as Financial Data and 10-K Sections.} Financial text analysis, surveyed in \cite{loughran2016, gentzkow2019}, has moved from dictionary counts \cite{loughran2011} and media sentiment \cite{tetlock2007} to topic models \cite{bao2014, dyer2017} and conference-call risk exposures \cite{hassan2019, hassan2023}. LLMs now extract corporate risks \cite{kim2023, dolphin2026}. Item 1A disclosures capture firm risk \cite{campbell2014}, but SEC rules assign distinct roles to Business (Item 1) and MD\&A (Item 7) \cite{sec10k}. Voluntary risk disclosures appeared in Business before the 2005 Item 1A mandate \cite{huang2022}, while forward-looking operational trends appear in MD\&A \cite{li2010}.

\textbf{Missing Data, Selection, and LLM Econometrics.} Whether inference can ignore the missing-data process depends on that process \cite{rubin1976, little2019}. For a bounded outcome such as a 0/1 indicator, Manski \cite{manski1989} derives bounds free of assumptions on selection. Selection models can correct for nonrandom data availability \cite{heckman1979} but should rest on justified exclusion restrictions and sensitivity analysis \cite{lennox2012}. Validation samples can correct bias in estimates that use machine predictions \cite{ludwig2026, egami2023, angelopoulos2023}. By contrast, a joint acceptance rule performs complete-case deletion \cite{little2019} inside the measurement instrument, before the researcher sees the data.

\textbf{Selective Prediction, Validation, and LLM Judges.} An acceptance rule is a reject option \cite{chow1970, geifman2017, wen2025}. In multilabel prediction, partial abstention \cite{nguyen2021} reports some labels and withholds others, as separate acceptance does. Data-engineering practice likewise distinguishes dropping invalid records from failing a whole update \cite{databricks2026}. Assertions in LLM pipelines can be synthesized to limit false failures \cite{shankar2024}. Grammar-constrained decoding can enforce output structure \cite{geng2023}, but a quotation can still fail verification. LLM judges exhibit biases \cite{zheng2023} and require validation against human judgments \cite{bavaresco2025}.

\section{Two Stages From Response to Variable}
\subsection{Stage 1: Acceptance Is Sample Selection}\label{sec:stage1def}
Each configuration $m$ reads a section in chunks and returns three outputs $t$: risk factors, current/past shocks, and future shocks. I call the two shock outputs the current/past field and the future field. Output $t$ of configuration $m$ for filing $f$'s Item 1A passes ($D_{fmt}=1$) when it passes its own checks in every chunk and after merging (Section~\ref{sec:corpus}). One such check is the \emph{three-array requirement}: every chunk response must be machine-readable and hold exactly the three arrays.

\emph{Separate acceptance} accepts each output that passes. \emph{Joint acceptance} accepts all three outputs if all three pass, and none otherwise. The original pipeline applied joint acceptance with one exception: it also accepted one configuration's section that fails the three-array requirement. I call the rule without that exception \emph{strict}, and Table~\ref{tab:stage1_accept} reports it. As the two differ by one section, I call both the joint rule.

Acceptance decides which outputs a researcher observes. For one configuration, output, and filing, $A=1$ if the rule in force accepts the output. Under separate acceptance $A=D_{fmt}$, and under strict joint acceptance $A=1$ exactly when all three outputs pass. $Y=1$ if the output would hold a retained record once accepted. A researcher observes $Y$ only when $A=1$. An output with $A=0$ is unresolved, whether rejected by the rule in force or never completed. It is missing, never zero. An accepted output with no retained record is a recorded zero ($Y=0$).

A study wants $\theta_m$, the share of filings with Item 1A text whose output would be positive if accepted. A researcher instead sees the complete-case share $q_m$, the positive share among accepted outputs, or the zero-imputed share $d_m$, which counts unresolved outputs as zeros. Let $r_m$ be the acceptance rate and $p_m$ the unobserved positive share among unresolved outputs. Given the stored responses, $Y$ and $\theta_m$ do not depend on the rule.

A common sample holds the filings accepted in all four configurations. It is the joint sample $J$ under strict joint acceptance and the separate sample $S_t$ for output $t$ under separate acceptance. $S$ is the current/past separate sample. Its entrants $E$ are the filings in $S$ but not in $J$. By construction, $S_t$ contains $J$ (Section~\ref{sec:corpus}).

\textbf{Proposition 1 (Acceptance as selection).} If \mbox{$0<r_m<1$},
\begin{equation}
\theta_m=r_mq_m+(1-r_m)\,p_m=d_m+(1-r_m)\,p_m. \label{eq:selection}
\end{equation}
\begin{itemize}
\item[(a)] \emph{Bias:} The zero-imputed share $d_m$ can only understate $\theta_m$. The complete-case share $q_m$ can err in either direction, depending on the unobserved $p_m$.
\item[(b)] \emph{Manski bounds}~\cite{manski1989}\emph{:} With no assumption on unresolved outputs, $\theta_m\in[d_m,\,d_m+1-r_m]$. Separate acceptance narrows this interval by the share that the strict joint rule rejects only because a companion output failed.
\item[(c)] \emph{Entrants:} With $\bar{Y}_X$ the mean of $Y$ over a set $X$ of $|X|$ filings, and $J$ and $E$ nonempty, $\bar{Y}_S-\bar{Y}_J=(|E|/|S|)\,(\bar{Y}_E-\bar{Y}_J)$.
\end{itemize}
Each part also holds within any subgroup of filings, such as a filing year, and for a common sample, where $A$ marks membership and $Y$ is any filing-level indicator. Biases and bound widths scale with the unresolved share $1-r_m$, so a yearly series can inherit trends in acceptance.

\subsection{Stage 2: Which Sections Are Read and Which Labels Count}\label{sec:stage2def}
Stage 2 holds acceptance fixed and varies the sections read and the labels counted. It fixes a configuration, a shock field (current/past or future), and the panel of filings in which all four configurations completed all three sections under the joint rule. The \emph{section rule} reads Item 1A alone or all three Items, 1, 1A, and 7. The \emph{label rule} sets the counted labels $\mathcal{C}$. The \emph{named rule} counts only the named event labels ($\mathcal{C}_N$), and the \emph{broad rule} also counts the residual \texttt{other\_macro} label ($\mathcal{C}_B$).

A filing is flagged when it has a counted label in the sections read. A filing is \emph{newly flagged} if Business or MD\&A has a counted label in that field and Item 1A has none. The increment $\Delta(\mathcal{C})$ is the share of newly flagged filings in the panel. I define \emph{added labels} as the distinct counted labels in that field a filing gains once all three Items are read. Table~\ref{tab:stage2_main} counts them as filing--label pairs. 

\textbf{Proposition 2 (Margins and membership).}
\begin{itemize}
\item[(a)] \emph{Hidden intensive margin:} $\Delta(\mathcal{C})$ ignores every label added to a filing already positive in Item 1A under $\mathcal{C}$.
\item[(b)] \emph{Non-nested added sets:} Each label rule can newly flag filings the other does not. A filing whose only Item 1A label is \texttt{other\_macro} is already positive under the broad rule, so gaining a named label newly flags it only under the named rule. A filing with no Item 1A label that gains only \texttt{other\_macro} is newly flagged only under the broad rule.
\end{itemize}
A binary exposure flag thus records only the extensive margin.

\section{Corpus, Configurations, and Replay Design}\label{sec:corpus}
My archive holds 15,600 original Form 10-K filings, filed 2000--2025, excluding real estate investment trusts and asset-backed issuers. I stratified the sample by filing year and primary-document size, a proxy for complexity. Each year has 600 filings, 200 in each of three size groups, drawn from a frame of 165,596 filings~\cite{gun2026bl}, so pooled shares describe the sample, not all filers. I identify firms by their SEC Central Index Key (CIK). The sample covers $9{,}603$ firms. Of the sampled filings, $3{,}202$ have no Item 1A section in my archive ($2{,}819$ filed 2000--2005). The $12{,}398$ available Item 1A sections comprise $1{,}271$ placeholders under 1,000 parsed characters (mostly exemption notices, plus contents lines and cross-references) and $11{,}127$ substantive sections.

Four LLM configurations processed every section. \texttt{Gemma26} (Gemma 4 26B-A4B~\cite{gemmateam2026gemma4}), \texttt{Gemma31} (Gemma 4 31B~\cite{gemmateam2026gemma4}), and \texttt{Qwen} (Qwen3.6 35B-A3B~\cite{qwen36_35b_a3b}) ran self-hosted with 8-bit quantization. \texttt{Nemotron} (Nemotron 3 Super 120B-A12B~\cite{nvidia_nemotron_3_2025}) ran on its provider's service.

I send sections in chunks of at most 110,000 characters, and 86\% of Item 1A sections fit in one chunk. Each chunk response returns three arrays: \texttt{risk\_factors} (one record per risk-factor heading in Item 1A, only explicit risks in Item 1, and only forward-looking risks in Item 7), \texttt{shocks\_current\_past}, and \texttt{shocks\_future}. The prompt asks for a verbatim quotation in every record. Each shock record maps to a closed taxonomy of 42 named historical events in six groups, plus \texttt{other\_macro}. The groups are trade and geopolitical (13 events, including Brexit 2016), macro-financial (8, global financial crisis 2008), regulatory (9, SOX 2002), technology and work (5, generative AI 2023), physical and health (6, COVID-19 2020), and labor (1, labor actions 2022).

The prompt defines a current/past shock as a macro or sector-wide shock that the filing says is affecting or has affected the firm. It reserves \texttt{other\_macro} for genuine macro or sector events absent from the taxonomy. A future record requires a specific event that the firm expects to affect it, even if the event has already occurred. Normalization folds typographic characters and trims each quotation to its first $60$ words. If the trimmed quotation does not match the source, normalization keeps its longest matching clause of at least eight words. Failing that, it replaces the quotation with the closest sufficiently similar sentence or $60$-word excerpt from a single paragraph. This single-paragraph repair requires a word-level similarity of at least 0.65. Normalization also removes duplicates and applies the event-year rule, which deletes records whose event starts after the filing year.

Validation checks required fields and quotation matches. It starts with risk factors and stops at the first failing record. These checks test form and provenance only. A companion audit of this pipeline found that normalization changed the stored text of $22.0\%$ of sampled \texttt{Gemma31} quotations and $59.5\%$ of sampled \texttt{Qwen} quotations~\cite{gun2026gtrex}. Tracing the same quotations through my normalization code, I find that the single-paragraph repair replaced $2.2\%$ of \texttt{Gemma31}'s quotations and $4.1\%$ of \texttt{Qwen}'s with source text. The other differences shorten or reformat the model's own words. The study behind that audit uses the same numerical cutoff of 0.65 and blocks replacements that, for example, change a number's sign or drop a checked negation. This \emph{guarded repair} replaced $805$ model quotations with source text~\cite{gun2026gtrex}. In that study, two outside readers independently judged whether each replacement in a random sample of $100$ of the $805$ kept the meaning of the model's quotation~\cite{gun2026gtrex}. Both readers judged that $74$ replacements kept the meaning and $7$ changed it, both marked $2$ unclear, and they disagreed on the other $17$. None of the labeled quotations came from this paper's extractions, and the readers judged none of my repairs, which use no guards. The three-array requirement is one of each output's own checks (Section~\mbox{\ref{sec:stage1def}}). Every other check, normalization step, and merge reads one array at a time, so an output that passes its own checks passes every check the joint rule would apply to it.

Stage 1 re-checks each chunk's latest stored response with that configuration's unchanged code, without a new model call. Replay reproduces every joint verdict of the original run. Separate acceptance checks the three-array requirement before testing each array with the other two empty, which leaves unchanged every array that the strict rule accepted. The joint rule also determined which response the pipeline stored: it asked a failed chunk again, for up to 15 answers in all, and kept only the latest response. Since responses varied across retries in all four configurations, a retry could overwrite a passing shock array. The re-check therefore describes stored responses, not a fresh run under separate acceptance, which could retain a passing shock array while retrying a failed output.

Item 1A became mandatory for fiscal years ending on or after December 1, 2005~\cite{sec2005}, and smaller reporting companies remain exempt~\cite{sec10k}. The sample still holds $781$ Item 1A sections from 2000--2005. In $761$ of them, my section splitter found a part titled ``Risk Factors'' with no Item 1A label.

Stage 2 uses 95\% percentile intervals from $2{,}000$ CIK-clustered bootstrap resamples. These intervals condition on the panel and omit model-call variation and shared calendar shocks. Stage 1 estimates and Stage 2 regressions use firm-clustered standard errors (SE). Audit shares use Wilson intervals on Kish effective sample sizes, with Newcombe intervals for differences. All intervals are pointwise and unadjusted for multiple comparisons.

\begin{table}[t]
\centering
\caption{Stage 1 Acceptance and Entrant Profile}
\label{tab:stage1_accept}
\scriptsize
\setlength{\tabcolsep}{2.0pt}
\begin{tabular*}{\columnwidth}{@{\extracolsep{\fill}}lrrrrrrr@{}}
\toprule
\multicolumn{8}{@{}l}{\emph{Panel A: Current/past acceptance, 12,398 Item 1A sections}} \\
& \multicolumn{2}{c}{Strict joint} & \multicolumn{2}{c}{Recovered} & \multicolumn{2}{c}{Separate} & \\
\cmidrule(lr){2-3} \cmidrule(lr){4-5} \cmidrule(lr){6-7}
Configuration & $n$ & \% & Pos. & Empty & $n$ & \% & Unres. \\
\midrule
Gemma26 & 11,417 & 92.1 & 683 & 216 & 12,316 & 99.3 & 82 \\
Gemma31 & 10,920 & 88.1 & 1,236 & 187 & 12,343 & 99.6 & 55 \\
Nemotron & 11,308 & 91.2 & 244 & 791 & 12,343 & 99.6 & 55 \\
Qwen & 12,195 & 98.4 & 66 & 129 & 12,390 & 99.9 & 8 \\
Common sample & 9,617 & 77.6 & \multicolumn{2}{c}{$+$2,608 enter} & 12,225 & 98.6 & 173 \\
\midrule
\end{tabular*}\par\nointerlineskip
\setlength{\tabcolsep}{1.8pt}
\begin{tabular*}{\columnwidth}{@{\extracolsep{\fill}}lrrrr@{\hspace{2\tabcolsep}\hspace{4pt}}rrrr@{}}
\multicolumn{9}{@{}l}{\makebox[0pt][l]{\emph{Panel B: Composition (\%) and current/past positive share (\%), substantive sections}}} \\
Group & $n$ & 2000--09 & $<$10k ch. & Distress & G26 & G31 & Nem. & Qwen \\
\midrule
Strict joint $J$ & 8,348 & 21.2 & 6.0 & 15.5 & 79.3 & 83.0 & 33.4 & 53.1 \\
Entrants $E$ & 2,607 & 40.2 & 1.7 & 17.4 & 79.2 & 85.8 & 27.0 & 47.6 \\
$S$ ($J$ plus $E$) & 10,955 & 25.8 & 5.0 & 16.0 & 79.3 & 83.7 & 31.9 & 51.8 \\
Change (points) & & $+$4.5 & $-$1.0 & $+$0.4 & 0.0 & $+$0.7 & $-$1.5 & $-$1.3 \\
\addlinespace[2pt]
Outside $S$ & 172 & 38.4 & 1.2 & 20.3 & --- & --- & --- & --- \\
All substantive & 11,127 & 26.0 & 4.9 & 16.0 & --- & --- & --- & --- \\
\bottomrule
\end{tabular*}
\par\vspace{2pt}
\parbox{\columnwidth}{\scriptsize The original run also accepted one placeholder \texttt{Nemotron} section (historical common sample of 9,618 sections). An output is recovered if it is unresolved under the strict joint rule but passes under the separate rule, with a shock record (Pos.) or without one (Empty). Unres.\ counts outputs still unresolved. Panel B omits the 1,271 placeholders. G26, G31, and Nem.\ abbreviate \texttt{Gemma26}, \texttt{Gemma31}, and \texttt{Nemotron}. Change is $S$ minus $J$ from unrounded shares. $<$10k ch.\ means under 10,000 characters.}
\end{table}

\section{Stage 1: What Joint Acceptance Throws Away}
Stage 1 varies one rule, acceptance, on the $12{,}398$ Item 1A sections. The stored responses and every other rule stay fixed. Stage 1 counts every retained current/past record, named or \texttt{other\_macro}. This is the broad label rule of Stage 2.

\subsection{Separate Acceptance Recovers Older, Longer Sections}
Joint acceptance keeps at least 88\% of available Item 1A sections in each configuration (Table~\ref{tab:stage1_accept}, Panel A). Across all four, it keeps only 78\%, because failures fall on different filings. Among the $11{,}127$ substantive sections it keeps $75.0\%$.

In Panel A, Skyworks' current/past output, rejected with its response in the Introduction, is a recovered positive. Separate acceptance recovers nearly all unresolved current/past outputs. The entrants lift the common sample to 98.6\%. Under separate acceptance, the future-shock common sample holds $12{,}310$ sections ($99.3\%$). Risk factors barely move, to $9{,}651$ sections ($77.8\%$), because nearly every jointly rejected section fails risk-factor acceptance in some configuration.

Joint acceptance selects heavily on length and era. Among substantive sections, the common sample keeps about 92\% of those under 10,000 characters and about three quarters of longer ones. It keeps only 61\% of sections filed before 2010, and its yearly share rises from 48\% in 2000 to 85\% in 2025 (Fig.~\ref{fig:stage1}(a)). Any study variable that moves with length or era can inherit this tilt toward short, recent sections. A zero-imputed series can understate the early years most, where acceptance is lowest (Proposition~1(a)), and can turn rising acceptance into rising incidence.

The entrants are older and longer than the joint sample (Table~\ref{tab:stage1_accept}, Panel B). Of the substantive entrants, $40.2\%$ were filed before 2010, against $21.2\%$ of the joint sample, and $1.7\%$ have fewer than 10,000 characters, against $6.0\%$. The $1{,}271$ placeholder sections, which Panel B excludes, yield no shocks and fall almost entirely in the joint sample.

The distress flag comes from a companion covenant extraction on the same filings by the same four models. It equals 1 when at least two of them report one of these items: a covenant breach, waiver, forbearance, acceleration or default notice, non-compliance, or a debt-linked going-concern doubt. Covenant violations can also occur at firms that are not in distress~\cite{dichev2002}, so the flag is a broad signal. I searched the main 10-K document and attached debt exhibits for covenant passages, excluding agreements incorporated by reference~\cite{gun2026bl}. For the $343$ of $12{,}398$ filings where this search yielded no covenant passage, I code the flag as zero. Within filing year, entrants are not detectably more likely than joint-sample sections to carry the flag.

By Proposition~1(c), the joint-sample and entrant rows ($J$ and $E$) exactly account for the positive-share shift in Panel B's Change row. The shift runs from $-1.5$ to $+0.7$ points. Its sign varies by configuration, so a study should report its own entrant contrast and should not assume the direction of the complete-case bias (Proposition~1(a)).

Most failing quotations differ from the source only in formatting or small omissions, so a failed match does not show a fabricated quotation~\cite{gun2026gtrex}. Among the risk-factor quotations that accompany the recovered positive current/past outputs, only $4.8\%$ failed both case- and whitespace-insensitive matching and single-paragraph repair, and most of these ($86\%$) keep the source's words in order. An exploratory \emph{list-aware check} matches contiguous words after removing list markers, punctuation, table-of-contents headers, page numbers, and back-to-top lines. Its yield is an in-sample best case. Under joint acceptance, it delivers about two thirds of the current/past common-sample gain from separate acceptance. With both changes, the risk-factor common sample reaches $92.1\%$.

\subsection{Does It Change Measured Exposure?}
\begin{table}[t]
\centering
\caption{Stage 1 Shock Levels and Regressions}
\label{tab:stage1_levels}
\scriptsize
\setlength{\tabcolsep}{3pt}
\begin{tabular*}{\columnwidth}{@{\extracolsep{\fill}}lrrr@{}}
\toprule
\multicolumn{4}{@{}l}{\emph{Panel A: Levels (\%) and gaps (points)}} \\
 & Joint $J$ & Separate $S$ & Change \\
\midrule
Manski bounds, common sample & [65.3, 90.3] & [86.3, 87.9] & width $-$23.4 \\
Manski bounds, all accepted outputs & [85.4, 90.3] & [87.8, 87.9] & width $-$4.7 \\
Any-configuration, all filing years & 87.1 & 87.7 & $+$0.6 \\
Any-configuration, 2009 filings & 79.5 & 83.5 & $+$4.0 \\
\quad including placeholders & 63.6 & 71.9 & $+$8.2 \\
\midrule
\multicolumn{4}{@{}l}{\emph{Panel B: Any-configuration shock flag, linear probability models (points)}} \\
Design: regressor & $J$ & $S$ & Change (SE) \\
\midrule
Year $+$ industry FE: log length & 16.94 & 16.74 & $-$0.20 (0.21) \\
\quad distress & 1.60 & 1.75 & $+$0.16 (0.42) \\
Year FE: log length & 16.79 & 16.12 & $-$0.67 (0.20) \\
\quad distress & 0.68 & 1.35 & $+$0.67 (0.42) \\
Firm $+$ year FE: log length & 15.27 & 15.59 & $+$0.32 (0.82) \\
\quad distress & 1.22 & 2.25 & $+$1.03 (1.01) \\
2008--10 only, year FE: distress & $-$0.47 & 2.07 & $+$2.54 (1.29) \\
\bottomrule
\end{tabular*}
\par\vspace{2pt}
\parbox{\columnwidth}{\scriptsize Substantive sections, except the placeholder row. Any-configuration equals 1 when at least one configuration retains a current/past record. Models with year FE alone include a financial-sector indicator. FE is fixed effects, SE is firm-clustered standard error, and changes use unrounded values.}
\end{table}

\begin{figure}[t]
\centering
\begin{tikzpicture}
\begin{groupplot}[
  group style={group size=1 by 2, vertical sep=6pt, x descriptions at=edge bottom},
  width=0.80\columnwidth, scale only axis,
  xmin=1999.5, xmax=2025.5,
  xtick={2000,2005,2010,2015,2020,2025},
  xticklabels={2000,2005,2010,2015,2020,2025},
  tick label style={font=\scriptsize}, label style={font=\scriptsize},
  ylabel style={align=center},
  ymajorgrids, grid style={gray!25}, axis line style={gray!60},
  axis x line*=bottom, axis y line*=left, every tick/.style={gray!60},
  legend cell align=left,
  legend style={inner sep=1pt, font=\scriptsize, draw=none, fill=white, fill opacity=0.9, text opacity=1,
                legend columns=2, /tikz/every even column/.append style={column sep=6pt}},
  clip=false]
\nextgroupplot[
  height=0.67in, ymin=40, ymax=104, ytick={40,60,80,100},
  ylabel={Common sample\\(\% of sections)},
  legend style={at={(0.995,0.03)}, anchor=south east}]
\fill[black!6] (axis cs:1999.5,40) rectangle (axis cs:2005.5,104);
\addplot[figblue, mark=*, mark size=0.9pt, line width=0.9pt] coordinates {(2000,97.6) (2001,95.7) (2002,94.1) (2003,98.6) (2004,97.4) (2005,96.7) (2006,97.4) (2007,97.4) (2008,99.1) (2009,98.5) (2010,97.7) (2011,99.4) (2012,99.2) (2013,99.6) (2014,99.0) (2015,98.2) (2016,98.4) (2017,99.0) (2018,97.9) (2019,98.5) (2020,98.0) (2021,99.4) (2022,98.9) (2023,99.3) (2024,98.2) (2025,98.9)};
\addlegendentry{Separate}
\addplot[figblue, mark=*, mark size=1.2pt, line width=0.6pt, mark options={fill=white}] coordinates {(2000,48.2) (2001,60.9) (2002,64.4) (2003,64.1) (2004,63.6) (2005,65.9) (2006,58.5) (2007,60.4) (2008,61.6) (2009,64.0) (2010,67.8) (2011,73.5) (2012,79.3) (2013,77.3) (2014,79.4) (2015,77.4) (2016,82.3) (2017,79.5) (2018,76.4) (2019,80.1) (2020,82.2) (2021,82.6) (2022,86.6) (2023,82.5) (2024,82.8) (2025,85.0)};
\addlegendentry{Joint}
\node[anchor=west, font=\scriptsize] at (axis cs:1999.6,82) {(a)};
\nextgroupplot[
  height=0.85in, ymin=-5.9, ymax=7.2, ytick={-4,-2,0,2,4,6},
  ylabel={Gap $S-J$\\(points)}, xlabel={Filing year},
  legend style={at={(0.995,0.01)}, anchor=south east}]
\fill[black!6] (axis cs:1999.5,-5.9) rectangle (axis cs:2005.5,7.2);
\addplot[black!55, line width=0.5pt, forget plot] coordinates {(1999.5,0) (2025.5,0)};
\addplot[figblue, mark=*, mark size=0.8pt, line width=0.9pt] coordinates {(2000,6.5432) (2001,5.8442) (2002,0.4030) (2003,-1.2088) (2004,3.8265) (2005,5.1136) (2006,5.1632) (2007,2.5796) (2008,3.6848) (2009,3.9988) (2010,2.2434) (2011,2.5454) (2012,0.9080) (2013,1.2289) (2014,0.2199) (2015,1.6069) (2016,0.3599) (2017,1.2226) (2018,-0.6629) (2019,0.7801) (2020,-0.2214) (2021,0.4985) (2022,0.2920) (2023,-0.0709) (2024,0.0297) (2025,0.0856)};
\addlegendentry{Gap $S-J$}
\addplot[figgray, dashed, mark=o, mark size=0.7pt, line width=0.8pt] coordinates {(2000,1.4815) (2001,1.4205) (2002,-3.0146) (2003,-4.1107) (2004,2.7143) (2005,2.8757) (2006,0.6211) (2007,-0.0647) (2008,0.7999) (2009,1.3756) (2010,0.4103) (2011,0.5959) (2012,-0.2439) (2013,0.3166) (2014,-0.1432) (2015,0.5145) (2016,0.0057) (2017,-0.0692) (2018,-1.4388) (2019,-0.0842) (2020,-0.4712) (2021,-0.0595) (2022,0.1323) (2023,-0.2492) (2024,-0.3668) (2025,-0.1669)};
\addlegendentry{After reweighting}
\addplot[only marks, mark=*, mark size=2pt, black, forget plot] coordinates {(2009,3.9988)};
\node[anchor=south west, font=\scriptsize, text=black] at (axis cs:2009.2,4.05) {2009: $+4.0$};
\node[anchor=west, font=\scriptsize] at (axis cs:1999.6,-5.0) {(b)};
\end{groupplot}
\end{tikzpicture}
\caption{Stage 1 by filing year, substantive Item 1A sections. (a) Share in the four-configuration current/past common sample under joint ($J$) and separate ($S$) acceptance. (b) Any-configuration shock gap ($S$ minus $J$) and gap left after reweighting $J$ to $S$'s length mix within year (cuts at 10,000, 30,000, 60,000, and 110,000 characters). Shaded: 2000--2005 pre-mandate years, each with under 200 sections.}
\label{fig:stage1}
\end{figure}
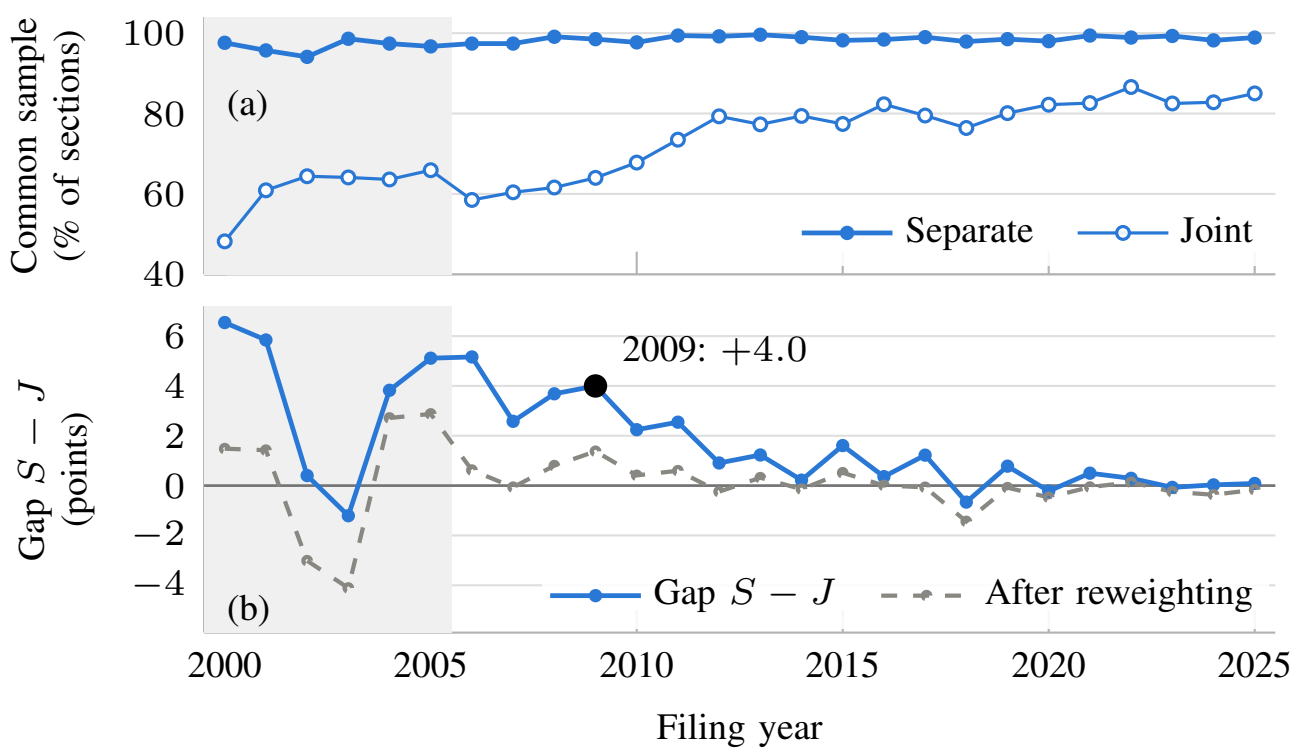

On substantive sections, separate acceptance narrows the logical bounds on the pooled any-configuration current/past shock rate (Table~\ref{tab:stage1_levels}, Panel A). Common-sample bounds count every section outside the common sample as unknown and narrow from 25 points wide to under 2. All-accepted-outputs bounds count a section as positive if any accepted output records a shock, and as empty only if all four outputs are accepted and empty. They narrow from 4.9 points wide to 0.1, computed from unrounded bounds. Both set unknown sections to 0 or 1 (Proposition~1(b)), so they are logical ranges on recorded machine outputs, not confidence intervals.

The yearly gap between the separate and joint samples in the any-configuration indicator is largest before 2012 and small afterward (Fig.~\ref{fig:stage1}(b)). In 2009 filings, which mostly report fiscal 2008, entrants make up about a third of the separate sample. They cite a shock more often than joint-sample sections. By Proposition~1(c), these two facts lift the indicator from 79.5\% to 83.5\% (Table~\ref{tab:stage1_levels}, Panel A), a gain of $4.0$ points (SE $1.1$). Counting named events only, the 2009 share rises from $70.1\%$ to $72.2\%$, a gain that is not detectable (SE $1.5$), and the pooled share falls from $65.5\%$ to $64.7\%$.

I reweighted the joint sample to the separate sample's counts in filing-year by length-group cells (length groups in Fig.~\ref{fig:stage1}). With all shock labels counted, reweighting on the shared year and length cells closes $88\%$ of the pooled gap and about two thirds of the 2009 any-configuration gap. Most of the level shift therefore reflects the tilt toward short, recent sections. The remaining third of the 2009 gap may reflect selection on characteristics I do not observe. Time-series designs should check acceptance and denominator composition by year, since either could change measured trends.

Pooled within-year and within-firm regression coefficients move by about one point or less when the acceptance rule changes (Table~\ref{tab:stage1_levels}, Panel B). Panel B regresses the any-configuration indicator on the natural log of section length and distress for sections with SIC codes. Industry effects use two-digit SIC codes. Across all years, the joint and separate samples contain 8,218 and 10,791 sections. Restricting the firm-effects models to repeat filers leaves 4,356 and 6,404. The SE of the coefficient change between the two samples allows for the joint sample being nested in the separate sample. Among the pooled designs, only the year-effects length coefficient shifts detectably. The year-effects design confined to 2008--2010 filings, with $956$ and $1{,}465$ sections, moves more and is less precise. The sign of its distress coefficient depends on the rule. Its change between rules, $+2.5$ points (SE $1.3$), sits at the edge of detectability, while neither estimate differs detectably from zero.

A cleaning rule can also manufacture zeros. Of the outputs accepted as empty under separate acceptance, all but about 2\% are explicit empty arrays. The event-year rule emptied the rest and deleted further records from accepted shock outputs. I audited $190$ deleted records with two separate blind runs of one LLM judge (Claude Opus 5.5) that agreed on every verdict. I drew them from $24$ strata of field, configuration, and event-year gap. After design weighting, the judge rated only about $6\%$ valid, mostly future-field expectations about rules enacted by the filing date but effective later. The judge accepted none of the $23$ sampled deletions from emptied current/past outputs. None of the $23$ came from \texttt{Gemma31}'s $98$ such outputs. Both decisions need logging: acceptance decides missing versus observed, and cleaning decides positive versus zero.

\begin{table}[t]
\centering
\caption{Stage 2 Increments, Rule Overlap, and Added Labels}
\label{tab:stage2_main}
\scriptsize
\begin{tabular*}{\columnwidth}{@{}l@{\extracolsep{\fill}}r@{\extracolsep{0pt}\hspace{8.5pt}}r@{\hspace{8.5pt}}r@{\hspace{12pt}}r@{\hspace{10pt}}r@{\hspace{7pt}}r@{\hspace{7pt}}r@{}}
\toprule
 & \multicolumn{3}{@{}c@{\hspace{12pt}}}{Filings flagged (\% of 8,987)} & & \multicolumn{3}{@{}c@{}}{Added named labels} \\
\cmidrule(r{12pt}){2-4} \cmidrule{6-8}
 & 1A, & & & IoU & & On 1A- & Other \\
Configuration & named & $\Delta(\mathcal{C}_N)$ & $\Delta(\mathcal{C}_B)$ & (\%) & Pairs & flagged (\%) & field (\%) \\
\midrule
\multicolumn{8}{@{}l@{}}{\emph{Current/past shocks}} \\
Gemma26 & 33.7 & 11.9 & 20.5 & 18.2 & 2,232 & 44.9 & 8.2 \\
Gemma31 & 49.7 & 15.0 & 15.8 & 33.5 & 3,731 & 54.1 & 4.5 \\
Nemotron & 27.1 & 12.0 & 15.3 & 72.8 & 1,512 & 25.3 & 1.3 \\
Qwen & 44.1 & 13.4 & 17.4 & 71.8 & 2,340 & 37.3 & 4.4 \\
\midrule
\multicolumn{8}{@{}l@{}}{\emph{Future shocks}} \\
Gemma26 & 17.4 & 9.0 & 28.5 & 15.6 & 1,195 & 27.4 & 44.5 \\
Gemma31 & 16.8 & 8.5 & 19.6 & 34.2 & 1,059 & 24.3 & 49.3 \\
Nemotron & 2.7 & 3.1 & 5.3 & 58.8 & 305 & 5.2 & 40.3 \\
Qwen & 11.3 & 14.7 & 25.1 & 56.2 & 1,717 & 14.3 & 45.2 \\
\bottomrule
\end{tabular*}
\par\vspace{2pt}
\parbox{\columnwidth}{\scriptsize 1A, named is the share of the 8,987 panel filings flagged from Item 1A alone under the named rule ($\mathcal{C}_N$). Each $\Delta$ is the share newly flagged when Items 1 and 7 are also read under $\mathcal{C}_N$ or broad $\mathcal{C}_B$. Intersection over union (IoU) is the share of filings newly flagged under either rule that both flag. Pairs counts distinct named filing--label pairs added beyond Item 1A. On 1A-flagged is the share of these pairs on filings where Item 1A already has another named label in the same field. Other field is the share whose label also appears in Item 1A's other shock field.}
\end{table}

\section{Stage 2: Reading More Sections and Counting More Labels}
Stage 2 varies two rules on the $8{,}987$ panel filings: the sections read and the labels counted. The stored responses and joint acceptance stay fixed. The joint rule selects the Stage 2 panel (Section~\mbox{\ref{sec:stage2def}}), and Item 1A is the binding section. Of the $12{,}324$ filings with all three source texts, $8{,}987$, from $6{,}242$ firms, pass joint acceptance in all three sections for all four configurations. Of these, $8{,}864$ have SIC codes. Under that rule, each configuration completes nearly all available Business and MD\&A sections. The panel leans toward recent years. Only $30$ of $600$ sampled filings from 2000 enter it, against most of those from 2025.

The panel contains $1{,}196$ filings ($13.3\%$) with a placeholder Item~1A. None has a retained Item 1A shock label, so a counted label in Business or MD\&A newly flags the filing in that field. Without them ($N=7{,}791$), the named-rule current/past increments of Table~\ref{tab:stage2_main} fall from $11.9$--$15.0$ points to $10.5$--$12.8$. The broad-rule increments fall from $15.3$--$20.5$ to $10.5$--$16.2$. Except where I note a placeholder-free check, the remaining Stage 2 tabulations and the Stage 2 audit of Section~\ref{sec:audit} use the full panel.

Common-panel estimates describe only the selected panel. On each configuration's own completed three-section panel, the named-rule increments differ from the common-panel values by at most about $0.2$ points for current/past and $0.4$ for future. On all filings with all three Items available, I let every output the joint rule left unresolved be positive or empty. This bounds each configuration's named-rule current/past increment within an interval at most $7.7$ points wide, with every endpoint between about 11 and 21 points. The measured changes from the two stages are not additive.

\subsection{More Sections Flag More Filings and Add Hidden Labels}
\begin{figure}[t]
\centering
\begin{tikzpicture}
\begin{axis}[
  width=0.96\columnwidth, height=1.80in,
  xmin=0, xmax=100, xtick={0,20,40,60,80,100},
  ymin=0.35, ymax=10.05,
  ytick={9,8,7,6,4,3,2,1},
  yticklabels={Gemma26, Gemma31, Nemotron, Qwen, Gemma26, Gemma31, Nemotron, Qwen},
  tick label style={font=\scriptsize}, label style={font=\scriptsize},
  ytick style={draw=none},
  xmajorgrids, grid style={gray!25}, axis line style={gray!60},
  axis x line*=bottom, axis y line*=left, every tick/.style={gray!60}, tick align=outside,
  xlabel={Panel filings with a counted shock label (\% of 8,987)},
  legend style={at={(axis cs:100.5,0.4)}, anchor=south east, legend columns=2, font=\fontsize{6}{7}\selectfont, inner sep=1pt, draw=none,
                fill=white, fill opacity=0.9, text opacity=1, row sep=-1pt,
                /tikz/every even column/.append style={column sep=5pt}},
  legend cell align=left,
  clip=false]
\addlegendimage{figblue, line width=1.2pt}
\addlegendentry{Named $\mathcal{C}_N$}
\addlegendimage{only marks, mark=o, mark size=1.6pt, black!70}
\addlegendentry{Item 1A only}
\addlegendimage{figorange, line width=1.2pt}
\addlegendentry{Broad $\mathcal{C}_B$}
\addlegendimage{only marks, mark=*, mark size=1.6pt, black!70}
\addlegendentry{Items 1, 1A, 7}
\node[font=\scriptsize\itshape, anchor=west, text=black!75] at (axis cs:0.5,9.72) {Current/past shocks};
\node[font=\scriptsize\itshape, anchor=west, text=black!75] at (axis cs:0.5,4.72) {Future shocks};
\addplot[figblue, line width=1.2pt, forget plot] coordinates {(33.7,9.17) (45.6,9.17)};
\addplot[figblue, line width=1.2pt, forget plot] coordinates {(49.7,8.17) (64.7,8.17)};
\addplot[figblue, line width=1.2pt, forget plot] coordinates {(27.1,7.17) (39.1,7.17)};
\addplot[figblue, line width=1.2pt, forget plot] coordinates {(44.1,6.17) (57.5,6.17)};
\addplot[figblue, line width=1.2pt, forget plot] coordinates {(17.4,4.17) (26.3,4.17)};
\addplot[figblue, line width=1.2pt, forget plot] coordinates {(16.8,3.17) (25.3,3.17)};
\addplot[figblue, line width=1.2pt, forget plot] coordinates {(2.7,2.17) (5.9,2.17)};
\addplot[figblue, line width=1.2pt, forget plot] coordinates {(11.3,1.17) (26.0,1.17)};
\addplot[only marks, mark=*, mark size=1.6pt, figblue, mark options={fill=white, line width=0.8pt}, forget plot]
  coordinates {(33.7,9.17) (49.7,8.17) (27.1,7.17) (44.1,6.17) (17.4,4.17) (16.8,3.17) (2.7,2.17) (11.3,1.17)};
\addplot[only marks, mark=*, mark size=1.6pt, figblue, forget plot]
  coordinates {(45.6,9.17) (64.7,8.17) (39.1,7.17) (57.5,6.17) (26.3,4.17) (25.3,3.17) (5.9,2.17) (26.0,1.17)};
\addplot[figorange, line width=1.2pt, forget plot] coordinates {(69.0,8.83) (89.4,8.83)};
\addplot[figorange, line width=1.2pt, forget plot] coordinates {(72.2,7.83) (88.0,7.83)};
\addplot[figorange, line width=1.2pt, forget plot] coordinates {(29.7,6.83) (44.9,6.83)};
\addplot[figorange, line width=1.2pt, forget plot] coordinates {(46.6,5.83) (63.9,5.83)};
\addplot[figorange, line width=1.2pt, forget plot] coordinates {(40.4,3.83) (68.9,3.83)};
\addplot[figorange, line width=1.2pt, forget plot] coordinates {(25.3,2.83) (44.9,2.83)};
\addplot[figorange, line width=1.2pt, forget plot] coordinates {(3.4,1.83) (8.6,1.83)};
\addplot[figorange, line width=1.2pt, forget plot] coordinates {(13.2,0.83) (38.3,0.83)};
\addplot[only marks, mark=*, mark size=1.6pt, figorange, mark options={fill=white, line width=0.8pt}, forget plot]
  coordinates {(69.0,8.83) (72.2,7.83) (29.7,6.83) (46.6,5.83) (40.4,3.83) (25.3,2.83) (3.4,1.83) (13.2,0.83)};
\addplot[only marks, mark=*, mark size=1.6pt, figorange, forget plot]
  coordinates {(89.4,8.83) (88.0,7.83) (44.9,6.83) (63.9,5.83) (68.9,3.83) (44.9,2.83) (8.6,1.83) (38.3,0.83)};
\node[font=\fontsize{6}{7}\selectfont, text=figblue, anchor=east] at (axis cs:32.3,9.17) {33.7};
\node[font=\fontsize{6}{7}\selectfont, text=figblue, anchor=west] at (axis cs:47.0,9.17) {45.6};
\node[font=\fontsize{6}{7}\selectfont, text=figorange, anchor=east] at (axis cs:67.6,8.83) {69.0};
\node[font=\fontsize{6}{7}\selectfont, text=figorange, anchor=west] at (axis cs:90.8,8.83) {89.4};
\end{axis}
\end{tikzpicture}
\caption{One set of stored responses, four variables. Share of the 8,987 panel filings with a counted shock label from Item 1A alone (open circles) to Items 1, 1A, and 7 (filled), under the named rule ($\mathcal{C}_N$, blue) or broad rule ($\mathcal{C}_B$, orange). Segment length equals $\Delta(\mathcal{C})$ in Table~\ref{tab:stage2_main}, up to rounding.}
\label{fig:rulespan}
\end{figure}
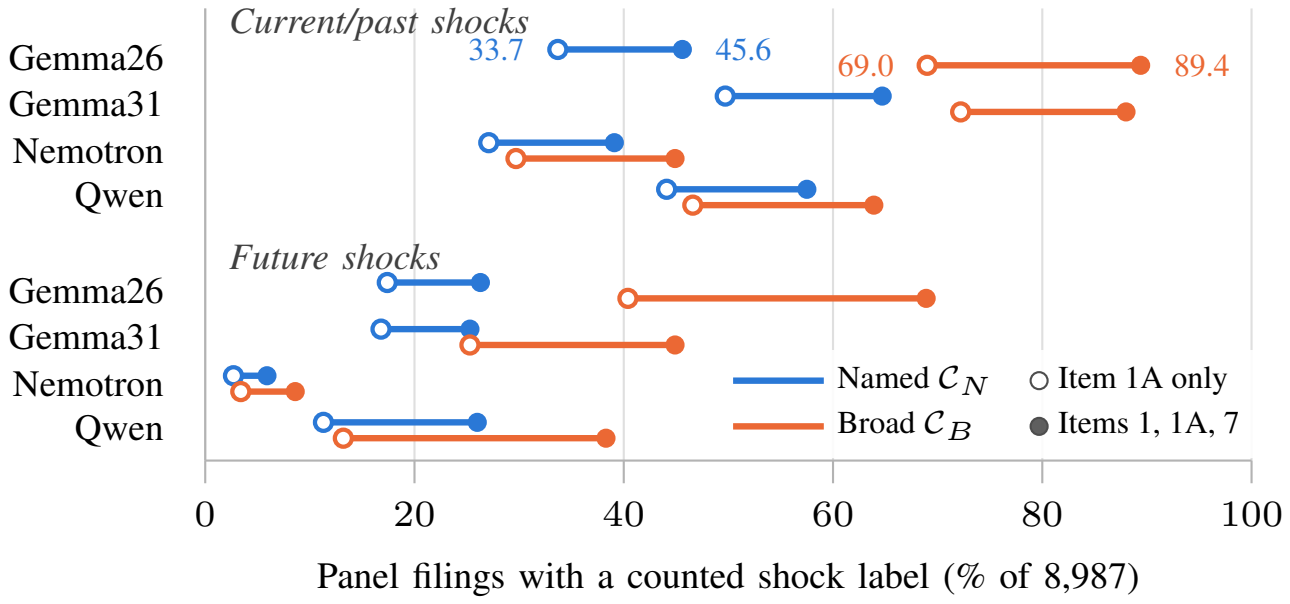

The same stored responses flag between $34\%$ and $89\%$ of panel filings for \texttt{Gemma26}'s current/past shocks, depending only on the section and label rules (Fig.~\ref{fig:rulespan}). Without placeholder filings, the range is $39\%$ to $95\%$. Reading Business and MD\&A alongside Risk Factors raises named-rule incidence in every configuration and field (Table~\ref{tab:stage2_main}, $N=8{,}987$). Every increment's CIK-clustered bootstrap 95\% interval lies within about one point of its estimate. No audit covers the added future records. Additions rarely repeat Item 1A text. Business or MD\&A quotations of at least eight words recur in Item 1A for under $4\%$ of added current/past labels under the named rule and under $8\%$ of future labels.

A binary flag registers none of the labels added to filings that Item 1A already flags. For \texttt{Gemma31}, these are more than half of its added current/past labels (Table~\ref{tab:stage2_main}, On 1A-flagged). Reporting a count of added labels beside the binary flag registers all of them. Between two fifths and half of added future labels are labels that the same filing's Item 1A records as current/past (Table~\ref{tab:stage2_main}, Other field).

Configurations agree only partly on the labels that Business and MD\&A add. About half to three quarters of each configuration's added named-rule current/past labels are also added by another configuration beyond Item 1A. Agreement across configurations may help screen additions for review, but only under the first judge of Section~\ref{sec:audit} does strict validity rise with the number of configurations that add a named label. Agreement cannot validate additions~\cite{gun2026bl}: all four read the same filing under one taxonomy, and a shared addition can repeat a shared misreading.

\subsection{The Label Rule Changes Which Filings Are Added}\label{sec:labelrule}
The label rule can change which filings are added even when it barely changes how many (Proposition~2(b)). For Unified Western Grocers, \texttt{Gemma31}'s Item 1A holds only an \texttt{other\_macro} current/past record, and its MD\&A adds the SOX label. The filing is therefore newly flagged only under the named rule. For \texttt{Gemma31}'s current/past shocks, the broad and named rules add nearly the same share of panel filings, under one point apart. Yet only a third of the filings added under either rule are added under both, an intersection over union (IoU) of $33.5\%$ (Table~\ref{tab:stage2_main}).

Across configurations, current/past IoU ranges from $18\%$ to $73\%$ (Table~\ref{tab:stage2_main}), or $11\%$ to $71\%$ without placeholder filings. \texttt{Gemma31}'s near-equal totals are not typical. Without placeholder filings, the named rule adds $2.2$ points more and the IoU is $23\%$. On the full panel, an event study that treats \texttt{Gemma31}'s newly flagged filings as exposed would reclassify $1{,}379$ filings while the group's size changes by only $75$.

\subsection{Does It Change a Regression?}
\begin{table}[t]
\centering
\caption{Stage 2 Distress Regressions Under Alternative Specifications}
\label{tab:stage2_reg}
\scriptsize
\setlength{\tabcolsep}{2.5pt}
\begin{tabular*}{\columnwidth}{@{\extracolsep{\fill}}lrrr@{}}
\toprule
 & Item 1A, & Items 1, 1A, 7, & Paired change \\
Specification & named & broad & (points) \\
\midrule
\multicolumn{4}{@{}l}{\emph{Panel A: Four-model distress flag (at least 2 of 4 models, 1,336 flagged)}} \\
Gemma26 & $-$3.0 (1.1) & $+$5.3 (1.4) & $+$8.2 (1.7) \\
Gemma31 & $-$1.1 (1.0) & $+$2.8 (1.4) & $+$3.8 (1.6) \\
Nemotron & $-$1.4 (1.1) & $+$0.6 (1.0) & $+$2.0 (1.1) \\
Qwen & $-$0.2 (1.0) & $+$0.3 (1.0) & $+$0.6 (0.9) \\
\midrule
\multicolumn{4}{@{}l}{\emph{Panel B: Gemma26, four-model flag, other samples and screens}} \\
(i) No placeholders, three lengths & $-$3.0 (1.1) & $+$4.5 (1.9) & $+$7.5 (2.1) \\
(ii) Screened \texttt{other\_macro} & $-$3.0 (1.1) & $+$2.1 (1.4) & $+$5.0 (1.7) \\
(i) and (ii) & $-$3.0 (1.1) & $+$0.1 (1.9) & $+$3.1 (2.1) \\
2008--2010 filings & $-$2.0 (4.0) & $+$4.2 (3.0) & $+$6.2 (4.8) \\
\midrule
\multicolumn{4}{@{}l}{\emph{Panel C: Gemma26, other distress flags (filings flagged)}} \\
At least 3 of 6 models (1,201) & $-$2.9 (1.0) & $+$5.7 (1.3) & $+$8.5 (1.6) \\
At least 2 of 6 models (1,651) & $-$3.2 (1.2) & $+$5.6 (1.6) & $+$8.8 (1.9) \\
Qwen3.8 Flash Next alone (1,771) & $-$4.3 (1.2) & $+$4.5 (1.7) & $+$8.8 (2.0) \\
GPT-OSS 120B alone (407) & $+$0.7 (0.7) & $+$2.2 (0.6) & $+$1.5 (0.9) \\
\bottomrule
\end{tabular*}
\par\vspace{2pt}
\parbox{\columnwidth}{\scriptsize Distress on current/past exposure (points, firm-clustered SEs in parentheses) for 8,864 SIC-coded filings with log Item 1A length and filing-year and industry effects unless noted. Paired change is broad three-section minus named Item 1A from unrounded values. Three lengths adds log Item 1 and 7 lengths (7,671 substantive filings). The screen drops \texttt{Gemma26} \texttt{other\_macro} records that mention terms such as liquidity, losses, default, or bankruptcy. The 2008--2010 row ($N=1{,}025$) omits industry effects. Six-model votes add GPT-OSS 120B and Qwen3.8 Flash Next.}
\end{table}
The rules also reach regression estimates (Tables~\ref{tab:headline} and~\ref{tab:stage2_reg}). I regress the distress flag on a current/past exposure indicator, with log Item 1A length and filing-year and industry effects. I use the $8{,}864$ jointly selected panel filings with SIC codes and change only the rules that define exposure.

With Item 1A and named labels, \texttt{Gemma26}'s coefficient is negative. With all three sections and broad labels, it is positive, and the change is detectable (Table~\ref{tab:stage2_reg}, Panel A). The three-section broad rule leaves only $948$ filings unexposed, so its coefficient compares most of the sample with a small remainder. Of those $948$ filings, $10.3\%$ carry the distress flag, against $13.7\%$ of the $4{,}011$ flagged through a named label and $17.6\%$ of the $3{,}905$ flagged only through \texttt{other\_macro}. This last group rests on the label family with lower strict validity in both audits (Section~\ref{sec:audit}, Table~\ref{tab:stage3_audit}). \texttt{Gemma31}'s change is smaller and also detectable. With industry effects, neither \texttt{Nemotron} nor \texttt{Qwen} has a detectable coefficient or change.

Changing only the labels moves \texttt{Gemma26}'s coefficient $3.0$ points (SE $1.4$) to near zero. Changing only the sections moves it $0.9$ points (SE $0.9$), leaving it negative (Table~\ref{tab:headline}). The sizable positive estimate appears when both rules change. The two-rule contrast is also detectable without the length control, with year effects only, and for an any-configuration indicator. In 2008--2010 filings, the change is not detectable (Table~\ref{tab:stage2_reg}, Panel B). Across all years, I also drop placeholders and control for the log lengths of Items 1 and 7, and \texttt{Gemma26}'s change remains detectable. Without placeholders, only $404$ of the $7{,}671$ regression filings remain unexposed under its broad three-section rule. \texttt{Gemma31}'s change is no longer detectable.

Part of the positive estimate may reflect firm-specific conditions filed under \texttt{other\_macro}. In the Stage 2 audit, each judge fails $17$ of $68$ \texttt{Gemma26} \texttt{other\_macro} records as not macro or sector-wide, against at most $2$ of $54$ named ones (unweighted). The Item 1A audit gives $16$ of $66$ against $1$ of $24$. Dropping \texttt{Gemma26}'s \texttt{other\_macro} records that mention terms such as liquidity, losses, default, or bankruptcy unflags $116$ filings and weakens the contrast (Table~\ref{tab:stage2_reg}, Panel B). After I also drop placeholders and control for section lengths, the change is no longer detectable. This screen removes only $3$ of the $17$ records that each judge fails: those on the firm's own losses, delisting notice, or bankruptcy. It also strips out genuine market events and bankruptcies of other firms. Both variables are machine readings of the same filing. The estimates are descriptive.

The distress flag also depends on which models vote. I ran the covenant extraction with two further models, GPT-OSS 120B and Qwen3.8 Flash Next, on the same passages with the same prompt~\cite{gun2026bl}. I used neither to extract shocks, so a flag from one of them alone shares no model with the exposure indicator. I add both to the vote and require at least three of the six models, which keeps the one-half share of the original rule. I repeat the regression with a second distress flag that requires at least two of the six models. The sign change holds under both six-model votes and under Qwen3.8 Flash Next's flag (Table~\ref{tab:stage2_reg}, Panel C). It does not hold under GPT-OSS 120B's flag, which is much sparser. Both of its coefficients are positive, and the change is not detectable.

\section{Are the Recovered and Added Records Valid?}\label{sec:audit}
\begin{table}[t]
\centering
\caption{Source-Level Precision Audits}
\label{tab:stage3_audit}
\scriptsize
\setlength{\tabcolsep}{2pt}
\begin{tabular*}{\columnwidth}{@{\extracolsep{\fill}}lrrrrrr@{}}
\toprule
 & \multicolumn{3}{c}{\makebox[0pt]{\emph{Panel A: Item 1A audit}}} & \multicolumn{3}{c}{\makebox[0pt]{\emph{Panel B: Stage 2 audit}}} \\
\cmidrule(lr){2-4} \cmidrule(l){5-7}
Current/past records (\%) & Recov. & Joint & Diff. & Added & Item 1A & Diff. \\
\midrule
Valid, strict: first judge & 28.2 & 31.9 & $-$3.8 & 44.3 & 33.3 & $+$11.0 \\
\quad second judge & 25.3 & 30.8 & $-$5.5 & 37.6 & 31.0 & $+$6.6 \\
Valid, lenient: first judge & 61.2 & 60.8 & $+$0.4 & 62.2 & 64.6 & $-$2.4 \\
\quad second judge & --- & --- & --- & 59.6 & 65.1 & $-$5.4 \\
Quotation supports event & 96.4 & 99.3 & $-$2.8 & 89.6 & 93.1 & $-$3.4 \\
\quad second judge & 96.4 & 97.6 & $-$1.3 & 76.2 & 91.5 & $-$15.3 \\
Macro or sector-wide & 84.9 & 84.8 & $+$0.1 & 93.5 & 88.9 & $+$4.7 \\
Taxonomy label correct & 61.2 & 61.1 & $+$0.2 & 63.2 & 65.6 & $-$2.4 \\
Realized firm effect & 53.1 & 53.5 & $-$0.4 & 67.6 & 54.5 & $+$13.1 \\
\midrule
\multicolumn{7}{@{}l}{\emph{Strictly valid by label family, unweighted}} \\
Named labels: first judge & 44.3 & 43.8 & & 58.3 & 45.8 & \\
\quad second judge & 27.8 & 37.5 & & 42.5 & 39.2 & \\
\texttt{other\_macro}: first judge & 16.9 & 21.4 & & 23.8 & 13.8 & \\
\quad second judge & 26.8 & 21.4 & & 31.3 & 18.8 & \\
\bottomrule
\end{tabular*}
\par\vspace{2pt}
\parbox{\columnwidth}{\scriptsize Panel A compares 150 recovered (Recov.) with 150 jointly accepted (Joint) records, and Panel B 200 added with 200 Item 1A records. Upper-block shares are weighted to the recovered or added mix (first judge Claude Opus 5.5 unless noted, second judge GPT-6.2 Sol). Lenient drops realized effect (post hoc in Panel A and omitted there for the second judge). Diff.\ is in points from unrounded shares. The lower block is unweighted. Not validated against human coders.}
\end{table}

Two blind audits compare the precision of recovered and added records with that of records already accepted. Precision is the share of records that a judge finds valid. Two recall checks ask what empty outputs miss. All four cover current/past records only, and the recall checks cover Item 1A alone.

An LLM judge (Claude Opus 5.5) reads each sampled record without being told its group. It rates the four criteria in Table~\ref{tab:stage3_audit} and gives an overall verdict, \emph{strict validity}. Two runs of the judge read every record, and a third run settles disagreements. The two runs agree on $95\%$ and $97\%$ of strict verdicts in the two audits. A second judge from another developer (GPT-6.2 Sol) read all $700$ audited records without being told the groups or the first judge's verdicts. It gives the same strict verdict on $86\%$ of records (chance-adjusted Cohen's $\kappa = 0.69$). Agreement is lower for recovered and added records ($84\%$ and $82\%$) than for their comparison groups ($90\%$ each).

The \emph{lenient} reading drops the realized-effect criterion and was post hoc in the Item~1A audit. The strict criteria are sharper than the extraction prompt in several places. For example, a possible effect (``could'', ``may'') fails the realized-effect criterion, and a persistent generic condition fails the label criterion for \texttt{other\_macro}. These sharper tests explain only a small part of the low strict shares. Given only the extraction prompt, a blind re-read by the first judge's model still fails $381$ of the $448$ records that the first judge's strict check fails across both audits, and it accepts $61$ of $62$ control records that passed the strict check. With that rate applied to the strict passes it did not cover, the unweighted share correct under the prompt is $45\%$, against $36\%$ under the strict check, and every comparison keeps its direction.

\emph{Recovered records (Panel A).} I compare $150$ Item 1A records that separate acceptance recovers with $150$ that the joint rule accepted. I match the groups on configuration, filing era, and section length, and weight both to the mix of recovered records. Weighting leaves effective sample sizes of $93$ and $101$. Neither judge detects a loss, but this does not establish equivalence. The first judge's strict difference is $-3.8$ points $[-16.4, +9.1]$. The interval rules out a loss above $16.4$ points but not one of $10$. For \texttt{Gemma31}, which supplies most recovered records, the difference is $-11$ points $[-28, +8]$. The second judge's difference is $-5.5$ points $[-17.8, +7.2]$, which also admits a loss of $10$. Recovered records are only $8\%$ of the current/past records accepted under separate acceptance. Even a $20$-point loss among them, beyond either judge's interval, would lower pooled precision by under two points.

\emph{Added records (Panel B).} In the Stage 2 panel, I compare $200$ records behind labels that Business and MD\&A add with $200$ Item 1A records. I match the groups on configuration, filing era, and label family, and weight both to the mix of added records. Neither judge detects a loss of strict validity in added records. The strict difference is $+11.0$ points $[+1.4, +20.3]$ under the first judge and $+6.6$ points $[-2.8, +15.7]$ under the second. Both intervals exclude a loss of $10$ points, but only the first excludes zero. Dropping the $35$ passages whose first line names the section, and so the group, moves neither difference by more than a point. Added records meet the realized-effect criterion more often under both judges. Their quotations support the event less often under the second judge, by $15$ points $[-22, -8]$. Among named labels, both judges rate added records lower on quotation support (by $9$ and $23$ points) and on the label (by $10$ and $20$ points), and all four intervals exclude zero. Neither lenient interval excludes a loss of $10$ points.

\emph{The residual label.} In every group of both audits, each judge finds \texttt{other\_macro} records strictly valid less often than named-label records (Table~\ref{tab:stage3_audit}, lower block), though the margin is only one point ($27.8\%$ against $26.8\%$) for the second judge's recovered records. Stage 2's broad rule therefore adds coverage from the less valid family. Audit precision does not identify recall or how errors covary with regression variables, so attenuation toward zero is not a safe default~\cite{ludwig2026}.

\emph{The judges.} Both judges rejected all $90$ records in which I planted a defect of known type, a check on clear-cut errors. On short constructed passages, both accepted all $20$ versions with a realized effect and rejected all $40$ with a possible effect or a hypothetical event. Of $60$ records that the first judge had accepted in both of its earlier runs, a fresh blind reading accepted $57$ under the first judge and $42$ under the second. This selection favors the first judge. I compare the two judge models with human readers on the guarded-repair reader study in Section~\ref{sec:corpus}. My coauthor on that study and I gave both models the readers' instructions and the same replacements. In one run each, the two models chose the same label on $86$ of the random $100$, and the two readers on $83$. On $6$ of the $86$, the two readers agreed on a different label, and in five their answer on whether the meaning changed was the opposite of the models'. The task differs from my audits, and none of the quotations comes from this paper's extractions, so these counts do not validate the audit verdicts. They show that two models can agree as often as two readers do and still share a label that both readers reject.

\emph{Recall.} First, a blind reader (Claude Opus 5.5) searched $40$ one-chunk Item 1A sections in which all four configurations returned no current/past shock. It found a qualifying statement in none of them ($95\%$ interval up to $9\%$). It found one in $25$ of $40$ comparison sections in which some configuration recorded a shock. It read each section twice, with a third reading for disagreements, and its first two readings agreed on $76$ of these $80$ sections. I then planted a passage from another filing in $18$ of the empty sections. One of the $18$ ends before it states the effect on the firm, so $17$ qualify. The first reader and a second reader (GPT-6.2 Sol) each found all $17$ and did not flag the one that does not qualify. Some planted passages did not fit the surrounding text. In the $22$ untouched sections, the first reader again flagged none. The second flagged $6$, all under the residual label. Five describe standing conditions or proposed rules, which the definition excludes. An empty output from all four configurations is therefore informative about named events, and less so about the residual label.

Second, for the $70$ named-label records that the first judge accepted in the Item 1A audit, I count how often each configuration returned the same label in the same section when another configuration supplied the sampled record. The unweighted share runs from $48\%$ for \texttt{Nemotron} ($24$ of $50$, 95\% interval $[35, 61]$) to $87\%$ for \texttt{Gemma31} ($39$ of $45$, $[74, 94]$). It omits events all four miss and events only the tested configuration finds, so it can overstate or understate recall. An empty output from one configuration can occur in a section where the judge accepts another configuration's named record.

\section{Checklist and Limitations}
Failure is not absence, and Risk Factors are not the whole filing. Table~\ref{tab:checklist} turns these findings into a release checklist. 

\begin{table}[t]
\centering
\caption{Release Checklist for LLM Extraction Datasets}
\label{tab:checklist}
\scriptsize
\setlength{\tabcolsep}{3pt}
\begin{tabular}{@{}l@{\hspace{5pt}}p{\dimexpr\columnwidth-5pt-0.5em\relax}@{}}
\toprule
1 & \textbf{Sources:} List expected filing--section pairs before joining outputs. Record source availability apart from model status. Keep sections with no records, and report shares with and without placeholder sections. \\
2 & \textbf{Missing vs.\ Zero:} Record chunk and section status. Store unresolved outputs as missing, never zero. Log cleaning deletions and keep raw responses with retries~\cite{gun2026gtrex}. \\
3 & \textbf{Acceptance:} Accept each output, section by section, only when its own checks pass in every chunk and after merging. Accept jointly only when the design requires agreement across outputs. Across sections, count a filing positive if any accepted section is positive, zero only if every section read is accepted and empty, and unresolved otherwise. \\
4 & \textbf{Samples and Bounds:} Report each common sample beside the available population. Give Manski bounds, compare excluded with retained sections within filing year, and reweight on pre-extraction characteristics. \\
5 & \textbf{Definitions:} State the acceptance rule, sections read, labels counted, and how current/past and future fields combine. Report label counts beside binary flags. \\
6 & \textbf{Screens and Audits:} Treat agreement across configurations as a screen, not validation. Audit precision blindly, comparing baseline with added records, and validate the audits against human coders. Read text with no output to gauge recall. \\
\bottomrule
\end{tabular}
\end{table}

Two LLM judges can share a misreading, so their agreement does not reveal an error both make. I did not audit the $200$ current/past outputs, in $173$ sections, that stay unresolved under separate acceptance.

In a companion study of the covenant extraction, I compared threshold flags with LLM readings of full filings, and most errors were omissions~\cite{gun2026bl}. Among $220$ sampled filings in which all four models recorded no threshold, those readings found a threshold in $16$, and every one traced to a shared pipeline rule. That audit did not cover the compliance statements or covenant events I use for the distress flag. If false-positive and false-negative rates do not vary by group and sum to less than one, misclassification pulls flag-rate differences toward zero. Error covarying with shocks or controls can bias either stage's regression coefficients upward or downward~\cite{ludwig2026}. Item 1A table-of-contents placeholders likely hide text the parser missed. The section splitter also misfiles text. In 3 of the 40 comparison sections of the recall check, the text filed as Item 1A is MD\&A text that follows a cross-reference to Risk Factors. A text scan finds the same opening, a cross-reference that ends in a closing quotation mark, in $75$ of the $11{,}127$ substantive sections. This scan misses unquoted cross-references, so $75$ is a lower bound.

 Each configuration bundles a model with its provider, quantization, and decoding, so contrasts between configurations do not isolate model size or architecture. With one run per configuration, I do not measure variation between runs. The magnitudes belong to this pipeline's quotation validator, retry design, taxonomy, and chunk size. Chunking explains little of the Stage 1 loss, which is $24.5\%$ among substantive sections that fit in one chunk and $27.8\%$ among split ones.

A pipeline that rejects a whole response on one failed check can discard a passing output. In a companion study of ten other models, $50.6\%$ of MD\&A quotations matched the source exactly, and allowing differences in whitespace, typography, and case raised the share to $95.6\%$~\cite{gun2026gtrex}. Except for Stage 2's binary section increments, the analyses are exploratory and retrospective, with no holdout. The models postdate the filings and may know how episodes unfolded, so the variables are not real-time measures.

\section{Conclusion}
A model and a prompt alone do not constitute a reproducible measurement procedure. Acceptance rules determine which observations enter the sample, while section and label rules determine what the variable captures. Holding stored LLM responses fixed, I show that joint acceptance favors shorter, more recent disclosures, while expanding section and label coverage increases measured current/past shock incidence from 34\% to 89\% in one configuration and reverses the sign of the covenant-distress regression coefficient across all four configurations, with detectable changes in two.

These findings have three implications for empirical research. First, failed extractions represent missing data, not evidence of absence. Second, broader coverage does not necessarily improve measurement quality, as both LLM judges assess residual-label records as less valid than named-label records. Third, stability in aggregate exposure increments does not imply stability in the identity of newly exposed firms, a distinction particularly relevant to empirical designs comparing exposed and unexposed firms.

The proposed release checklist provides a framework for documenting these methodological choices. Just as researchers report sample restrictions, variable definitions, and data-cleaning procedures, they should report acceptance rules, source sections, and counted labels alongside models and prompts. Reproducibility in LLM-based empirical research requires reproducing not only the model's responses but also the procedure that converts those responses into data.

\section*{Acknowledgment}
I am forever grateful to Azize Kadıoğlu for her guidance and support.

Large language models are central to the empirical analysis and are described in the sections that use their outputs. Models from Alibaba, Anthropic, DeepSeek, Google, and OpenAI also assisted with research design, coding, and checks, and, along with the Finference Prose model~\cite{gun2026prose}, helped draft and revise the text.

\end{document}